\documentclass[letterpaper]{article}
\usepackage[letterpaper]{geometry}

\usepackage[utf8]{inputenc}
\usepackage[T1]{fontenc} 
\usepackage[cjk]{kotex}

\usepackage{amsmath}
\usepackage{graphicx}

\usepackage{url}
\usepackage{xcolor}
\usepackage{latexsym}

\usepackage{hyperref}
\definecolor{darkblue}{rgb}{0, 0, 0.5}
\hypersetup{colorlinks=true,citecolor=darkblue, linkcolor=darkblue, urlcolor=darkblue}
\usepackage{tabularx}
\usepackage{datetime}
\usepackage{arydshln}

\usepackage{setspace}
\usepackage{lineno}

\usepackage[authoryear,round,longnamesfirst]{natbib}
\usepackage[ruled,vlined]{algorithm2e}
\usepackage[linguistics]{forest} 
\usepackage{synttree}
\usepackage{subcaption}
\usepackage{tikz-dependency}

\usepackage{gb4e}

\title{\textbf{{Korean Named Entity Recognition Based on Language-Specific Features}}}
\author{
Yige Chen$^{1\dagger}$, KyungTae Lim$^{2\dagger}$, and {Jungyeul Park}$^{3}$\\
$^{1}$The Chinese University of Hong Kong, Hong Kong\\
$^{2}$Seoul National University of Science and Technology, Seoul, South Korea\\
$^{3}$The University of British Columbia, Vancouver, BC, Canada \\
\footnotesize{$^{\dagger}$Yige Chen and KyungTae Lim contributed equally.} }
\date{ 
}

\begin{document}

\maketitle

\begin{abstract}
{In the paper, we propose a novel way of improving named entity recognition in the Korean language using its language-specific features.}
While the field of named entity recognition has been studied extensively in recent years, the mechanism of efficiently recognizing named entities in Korean has hardly been explored. This is because the Korean language has distinct linguistic properties that prevent models from achieving their best performances. Therefore, an annotation scheme for {Korean corpora} by adopting the CoNLL-U format, which decomposes Korean words into morphemes and reduces the ambiguity of named entities in the original segmentation that may contain functional morphemes such as postpositions and particles, is proposed herein. We investigate how the named entity tags are best represented in this morpheme-based scheme and implement an algorithm to convert word-based {and syllable-based Korean corpora} with named entities into the proposed morpheme-based format. Analyses of the results of {statistical and neural} models reveal that the proposed morpheme-based format is feasible, and the {varied} performances of the models under the influence of various additional language-specific features are demonstrated. Extrinsic conditions were also considered to observe the variance of the performances of the proposed models, given different types of data, including the original segmentation and different types of tagging formats.
\end{abstract}
\doublespacing

\section{Introduction} \label{intro}

Named entity recognition (NER) is a subfield of information extraction that searches for and extracts named entities (NE) such as locations, person names, organizations, and numbers from {texts} \citep{tjongkimsang-demeulder:2003:CoNLL}. 
With the rapid {development} in the field of machine learning, recent works on NER rely significantly on machine learning methods instead of grammar-based symbolic rules, with a particular focus on neural network algorithms. 
Numerous studies have been conducted on NER {concerning several natural languages, including} 
\citet*{carreras-EtAl:2003:EACL} for Catalan, 
\citet*{li-mccallum:2003} for Hindi,
\citet*{nouvel-EtAl:2013:TALN} {and \citet{park:2018:TALN:NER}} for French, 
and \citet{benikova-biemann-reznicek:2014:LREC,benikova-EtAl:2015} for German,
\citet{ahmadi-moradi:2015} for Persian. {There have also been language-independent multilingual models proposed} \citep{tjongkimsang-demeulder:2003:CoNLL,nothman-EtAl:2013,hahm-EtAl:2014:LREC}.
However, NER for the Korean language has not been explored thoroughly in previous research. This is partially because NE detection in Korean is difficult,\label{R3Q1} {as the language} does not utilize specific features {existing in some other languages} such as capitalization to emphasize NEs \citep{chung-hwang-jang:2003}. 
There have been studies and attempts on this topic, {one of the examples being an HMM-based co-trained} model in which co-training boosts the performance of HMM-based NE recognition \citep{chung-hwang-jang:2003}. 
Recently, neural methods have been applied to the NER tasks of Korean, which include Bi-LSTM-CRF with masked self-attention \citep{JIN2021101134}, as well as BERT \citep{kim-lee:2020}.
Such approaches require annotated corpora of high quality, which are usually difficult to source for Korean. 

Although gold standard corpora, such as CoNLL-03 \citep{tjongkimsang-demeulder:2003:CoNLL}, are still not available for the Korean language, {it is probable} that by constructing silver standard corpora with automatic approaches, both NE annotation and NER modeling can be {handled} decently. A silver standard NE corpus {in which NE annotations are completed automatically} has been constructed based on Wikipedia and DBpedia Ontology including Korean \citep{hahm-EtAl:2014:LREC}.
This high-quality corpus outperforms the manually annotated corpus. 
{On the other hand,} there have been new attempts that build a Korean NER open-source dataset based on the Korean Language Understanding Evaluation (KLUE) project \citep{park-EtAl:2021:KLUE}. 
KLUE's NER corpus is built mainly with the WIKITREE corpus\footnote{\url{https://www.wikitree.co.kr/}}, which is a body of news articles containing several types of entities.

Due to the linguistic features of the NE in Korean, conventional \textit{eojeol}-based segmentation, which makes use of whitespaces to separate phrases, does not produce ideal results in NER tasks. 
Most language processing systems and corpora developed for Korean {use} \textit{eojeol} delimited by whitespaces in a sentence as the fundamental unit of {text} analysis in Korean. This is partially because the Sejong corpus, the most widely-used corpus for Korean, employs \textit{eojeol} as the basic unit. The rationale of \textit{eojeol}-based processing is simply {treating} the words as they are in the surface form.
It is necessary for a better format and a better annotation scheme for the Korean language to be adapted. In particular, a word should be split into its morphemes.

To capture the language-specific features {in Korean and utilize them to boost the performances of NER models}, {we propose a new morpheme-based scheme for Korean NER corpora that handles NE tags on the morpheme level based on the CoNLL-U format designed for Korean, as in \citet{park-tyers:2019:LAW}. We also present an algorithm that converts the conventional Korean NER corpora into the {morpheme-based} CoNLL-U format, which includes not only NEs but also the morpheme-level information based on the morphological segmentation.}
The contributions of this study for Korean NER are as follows\label{main-conteribution}:
(1) An algorithm is implemented {in this study} to convert the Korean NER corpora to the proposed morpheme-based CoNLL-U format. We have investigated the best method to represent NE tags in the sentence along with their linguistic properties {and therefore developed the conversion algorithm with sufficient rationales}. 
(2) The proposed Korean NER models {in the paper} are distinct from other systems since our neural system is simultaneously trained for part-of-speech (POS) tagging and NER with a unified, continuous representation. This approach is beneficial {as it captures} complex syntactic information between NE and POS. {This is only possible with a scheme that contains additional linguistic information such as POS.} 
(3) The proposed morpheme-based scheme for NE tags provides a satisfying performance based on the automatically predicted linguistic features. Furthermore, we thoroughly investigate various POS types, {including} the language-specific XPOS and the universal UPOS \citep{petrov-das-mcdonald:2012:LREC}, and determine the type that has the most effect. We {demonstrate and account for the fact that} the proposed BERT-{based} system with linguistic features yields better results over those in which such linguistic features are not used.

{In the paper, we summarize previous work on Korean NER in Section \ref{previous-work}, and present a linguistic description of named entities in Korean in Section \ref{linguistics}, focusing on the features that affect the distribution of NE tags. We further provide a methodology for representing named entities in Korean in Section \ref{conllu}.}
Subsequently, we {introduce} morphologically enhanced Korean NER models {which utilize} the predicted UPOS and XPOS features in Section \ref{learning}. 
To evaluate the effect of the proposed method, we test the proposed morpheme-based CoNLL-U {data with NE tags} using three {different} models, namely CRF, RNN-based, and BERT-based models. 
We provide detailed experimental results of NER for Korean, {along with an} error analysis (Sections \ref{experiments} and Section \ref{results}). 
Finally, we present the conclusion (Section \ref{conclusion}).

\section{Linguistic description of NEs {in Korean}} \label{linguistics}

\subsection{Korean linguistics redux}\label{linguistics-redux}

{Korean is considered to be an agglutinative language in which functional morphemes are attached to the stem of the word as suffixes. The language possesses a subject-object-verb (SOV) word order, which means the object precedes the verb in a sentence. This is different from English grammar where the object succeeds the verb in a sentence. Examples from \citet{park-kim:2023} are presented as follows.}

{\begin{exe}
\ex \label{john-aux}
        \begin{xlist}
    \ex   \label{john-aux-obj}  \glll
    존이  다리도 건넌다 \\
    \textit{jon-i} \textit{dali-do}  \textit{geonneo-n-da} \\
\text{John-\textsc{nom}} \text{bridge-\textsc{also}}  \text{cross-\textsc{pres}-\textsc{dcl}}\\
\trans `John crosses the bridge, too'
    \ex \label{john-aux-obj-var}
    \glll
    다리도  존이 건넌다 \\
    \textit{dali-do} \textit{jon-i}   \textit{geonneo-n-da} \\
\text{bridge-\textsc{also}} \text{John-\textsc{nom}}  \text{cross-\textsc{pres}-\textsc{dcl}}\\
\trans `John crosses the bridge, too'
    \ex \label{john-aux-null}
    \glll 
    존 다리를 건넌다 \\
    \textit{jon} \textit{dali-leul}   \textit{geonneo-n-da} \\
    \text{John} \text{bridge-\textsc{acc}} \text{cross-\textsc{pres}-\textsc{dcl}}\\
    \trans `John crosses the bridge' 
\end{xlist}
    \end{exe}
}

{Two properties of the Korean language are observed. 
While Korean generally follows the SOV word order as in \eqref{john-aux-obj}, scrambling may occur from time to time which shifts the order of the subject and the object. 
Moreover, the postpositions in Korean, such as the nominative marker \textit{-i} in \eqref{john-aux-obj} and \eqref{john-aux-obj-var}, follow the stems to construct words, but there are also words in Korean that take no postposition, such as the subject \textit{jon} (`John') as presented in \eqref{john-aux-null}.}

{Another notable property of Korean is its natural segmentation, namely \textit{eojeol}. An \textit{eojeol} is a segmentation unit separated by space in Korean.  
Given that Korean is an agglutinative language, joining content and functional morphemes of words (\textit{eojeols}) is very productive, and the number of their combinations is exponential. We can treat a given noun or verb as a stem (also content) followed by several functional morphemes in Korean. Some of these morphemes can, sometimes, be assigned its syntactic category.
Let us consider the sentence in~\eqref{ungaro}.
\textit{Unggaro} (`Ungaro') is a content morpheme (a proper noun) and a postposition \textit{-ga} (nominative) is a functional morpheme. 
They form together a single \textit{eojeol} (or word) \textit{unggaro-ga} (`Ungaro+\textsc{nom}'). 
The nominative case markers \textit{-ga} or \textit{-i} may vary depending on the previous letter -- vowel or consonant. 
A predicate \textit{naseo-eoss-da} also consists of the content morpheme \textit{naseo} (`become') and its functional morphemes, \textit{-eoss} (`\textsc{past}') and \textit{-da} (`\textsc{decl}'), respectively.}

\begin{exe}
 \ex \label{ungaro}
 \glll
 {프랑스의}    {세계적인}    의상    디자이너    엠마누엘    {웅가로가}    실내    장식용    {직물}    디자이너로    {나섰다.} \\ 
 \textit{peurangseu-ui} {\textit{segye-jeok-i-n}} \textit{uisang} \textit{dijaineo} \textit{emmanuel} {\textit{unggaro-ga}} \textit{silnae} \textit{jangsik-yong   } \textit{jikmul   } \textit{dijaineo-ro} \textit{{naseo-eoss-da}.}\\ 
 France-\textsc{gen} world~class-\textsc{rel} fashion designer Emanuel Ungaro-\textsc{nom} interior    decoration textile designer-\textsc{ajt} become-\textsc{past}-\textsc{decl}\\
 \trans 'The world-class French fashion designer Emanuel Ungaro became an interior textile designer.' \\
\end{exe}


\subsection{NEs in Korean}

Generally, the types of NE in Korean do not differ significantly from those in other languages. NE in Korean consists of person names (PER), locations (LOC), organizations (ORG), numbers (NUM), dates (DAT), and other miscellaneous entities. Some examples of NE in Korean are as follows:  

\begin{center}
\begin{itemize}
    \item PER: 이창동 \textit{ichangdong} (`Lee Chang-dong', a South Korean film director)
    \item LOC: 제주 \textit{jeju} (`Jeju Island')
    \item ORG: 서울대학교 \textit{seouldaehaggyo} (`Seoul National University')
\end{itemize}
\end{center}

\begin{figure}
\centering
\includegraphics[width=0.9\textwidth]{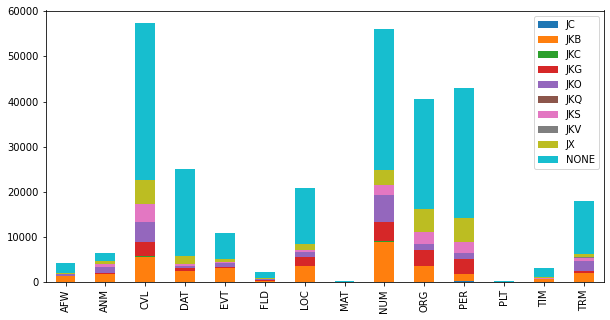}
\caption{Distribution of each type of postposition/particle after NEs ({NER data from NAVER}). {The terms and notations in the figure are described in Table \ref{ne-xpos-captions-table}.}
}\label{ne-xpos}
\end{figure}

While NEs can appear in any part of the sentence, they typically occur with certain distributions. Figure \ref{ne-xpos} presents the overall distribution of the postpositions after NEs, whereas Table \ref{percent-nepos} lists the percentage of the postpositional markers with respect to three types of typical NEs, namely PER, ORG, and LOC, based on {NAVER's Korean NER dataset}.\footnote{The source of {Figure~\ref{ne-xpos}, Table \ref{ne-xpos-captions-table}, and Table~\ref{percent-nepos}} is the morpheme-based CoNLL-U data with NE annotations we generate for this study using {the NER data from NAVER}, which will be introduced in detail in Section \ref{experiments}.}
For instance, a PER or an ORG is often followed by a topic marker {\textit{eun/neun}} (JX), a subject marker \textit{-i/-ga} (JKS), or an object marker \textit{-eul/-leul} (JKO).\label{R3Q12}
As detailed in Table~\ref{percent-nepos}, the percentages of JX, JKS, and JKO after ORG and PER are sufficiently high for us to make this conclusion, considering that in several cases, NEs are followed by {other grammatical categories (e.g., noun phrases), which takes up} high percentages {(58.92\% for LOC, 60.13\% for ORG, and 66.72\% for PER)}.
Moreover, a PER or an ORG tends to occur at the front of a sentence, which serves as its subject. Higher percentages of occurrences of JX and JKS than JKO after ORG and PER prove this characteristic. 
Meanwhile, a LOC is typically followed by an adverbial marker (JKB) such as \textit{-e} (locative `to'), \textit{-eseo} (`from') or {-(\textit{eu})\textit{ro}} (`to'), because the percentage of JKB after LOC is considerably {higher}.

\begin{table}
\caption{{XPOS (Sejong tag set) introduced in the morpheme-based CoNLL-U data converted from NAVER's NER corpus and its type of named entities}.}\label{ne-xpos-captions-table}
{\begin{minipage}{25pc}
\begin{tabular} {p{0.1\textwidth} p{0.4\textwidth} p{0.1\textwidth} p{0.5\textwidth}}    \hline
\multicolumn{2}{c}{XPOS}  & \multicolumn{2}{c}{Type of NE} \\ \hline 
NONE & non-postposition & AFW & artificially made articles: \textit{e.g.} 여객기 \textit{yeogaeggi} (`airliner') \\ \hdashline
JX & topic marker (은$|$는 \textit{eun$|$neun}) & ANM & animal: \textit{e.g.} 플랑크톤 \textit{peullangkeuton} (`plankton') \\ \hdashline
JKV & vocative marker (아$|$야 \textit{a$|$ya})  & CVL & civilization or culture related: \textit{e.g.} 홍보대사 \textit{hongbodaesa} (`ambassador') \\ \hdashline
JKS  & nominative marker (이$|$가 \textit{i$|$ga}) & DAT & date: \textit{e.g.} 주말 \textit{jumal} (`weekend') \\ \hdashline
JKQ  & postposition for quotations (고 \textit{go}) & EVT & event: \textit{e.g.} 챔피언스리그 \textit{chaempieonseuligeu} (`Champions League') \\ \hdashline
JKO & accusative marker (을$|$를 \textit{eul$|$leul})  & FLD & academic fields, theories, laws, and technologies: \textit{e.g.} 음성 인식 \textit{eumseong insig} (`Speech Recognition')\\ \hdashline
JKG  & genitive marker (의 \textit{ui}) & LOC & location: \textit{e.g.} 아메리카 \textit{amelika} (`America') \\ \hdashline
JKC & postposition for complements (이$|$가 \textit{i$|$ga}, same as of JKS)& MAT & metals, rocks, and chemicals material: \textit{e.g.} 칼슘 \textit{kalsyum} (`calcium') \\ \hdashline
JKB  & postposition for adverbs (에$|$에서 \textit{e$|$eseo}) & NUM & number: \textit{e.g.} 세번 \textit{sebeon} (`third time') \\ \hdashline
JC  & postposition for conjunctions (와$|$과 \textit{wa$|$gwa})  & ORG & organization: \textit{e.g.} 서울대학교 \textit{seouldaehaggyo} (`Seoul National University') \\ \hdashline
 & & PER & person: \textit{e.g.} 세종대왕 \textit{sejongdaewang} (`Sejong the Great')\\ \hdashline
 & & PLT & plant: \textit{e.g.} 포도나무 \textit{podonamu} (`vine')\\ \hdashline
 & & TIM & time: \textit{e.g.} 한시간 \textit{hansigan} (`one hour')\\ \hdashline
 & & TRM & term: \textit{e.g.} 납 중독 \textit{nab jungdog} (`lead poisoning')\\ \hline
\end{tabular}
  \end{minipage}}
\end{table}

\begin{table}
\caption{Percentage (\%) of occurrences of JKB/JKO/JKS/JX {among all types of parts-of-speech} after LOC/ORG/PER ({NER data from NAVER}).}\label{percent-nepos}
{\begin{minipage}{25pc}
\begin{tabular}{c|cccc}
\hline
\  & JKB  & JKO  & JKS  & JX   \\ \hline
LOC & {16.72} & {4.92} & 2.59 & {6.51}  \\ \hdashline
ORG & {8.61}  & {3.38} & {6.30} & {12.49} \\ \hdashline
PER & 3.50  & 2.94 & {5.49} & {12.72} \\ \hline
\end{tabular}
  \end{minipage}}

\end{table}

As mentioned in Section \ref{intro}, NE in the Korean language is considered to be difficult to handle compared to {Latin} alphabet-based languages such as English and French. This is mainly because of the linguistic features of NE in Korean.  
The Korean language has a unique phonetic-based writing system known as \textit{hangul}, which does not differentiate lowercase and uppercase. Nor does it possess any other special forms or emphasis regarding NEs. Consequently, an NE in Korean, through its surface form, makes no difference from a non-NE word. Because the Korean writing system is only based on its phonological form owing to the lack of marking on NE, it also creates type ambiguities when it comes to NER.

Another feature of NE in Korean is that the natural segmentation of words or morphemes does not always represent NE pieces. The segmentation of Korean is based on \textit{eojeol}, and each \textit{eojeol} contains not only a specific phrase {as a content morpheme} but also postpositions or particles that are attached to the phrase {as functional morphemes}, which is typical in agglutinative languages. 
{Normally}, Korean NEs only consist of noun phrases and numerical digits. But under special circumstances, a particle or a postposition could be part of the NE, such as {the} genitive case maker \textit{-ui}.
Moreover, an NE in Korean does not always consist of only one \textit{eojeol}. In Korean, location and organization can be decomposed into several \textit{eojeols} \citep{yun:2007}, which implies that there is no one-to-one correspondence with respect to {a Korean NE} and its \textit{eojeol} segment.
 
Therefore, the main difficulties in Korean NEs are as follows: (1) NEs in Korean are not marked or emphasized, which requires other features from adjacent words or the entire sentence to be used; (2) In a particular \textit{eojeol} segment, one needs to specify the parts of this segmentation that belong to the NE, and within several segments, one needs to determine the segments {that correspond to} the NE.

\section{Previous work} \label{previous-work}

{
{Korean NER is usually considered to be a sequence-labeling problem that assigns tags based on the boundary of the morphemes or the \textit{eojeols} (words or phrases separated by whitespaces in Korean).}
Thus, two {major} approaches, namely morpheme- and \textit{eojeol}-based NER have been proposed {depending on their corresponding }target boundary. 
{The morpheme-based approach separates the word into a sequence of morphemes for detection of the named entity, and accordingly, the recognized named entity is the sequence of morphemes itself. The \textit{eojeol}-based approach, on the other hand, uses the word (or \textit{eojeol}) as the basic unit for detection of the named entity, and the recognized entity may contain extra morphemes such as functional morphemes (e.g. the case marker) which should be excluded from the named entity. However, this post-processing task has been ignored in the \textit{eojeol}-based system.}
{Apart from the two major approaches described above, some Korean NER datasets also adopt the syllable-based annotation approach which splits Korean texts into syllables and assigns NE tags directly on the syllables.}
Although the target boundaries of NER tasks between {these} approaches differ considerably, {the statistical and neural systems for performing NER remain identical for both approaches as they both rely on the sequence-labeling algorithm.} Figure~\ref{korean-ner-data-examples} {illustrates} various annotation approaches for NEs in Korean, including \textit{eojeol}, morpheme, and syllable.

\begin{figure}
    \centering
{\small 
\begin{tabular} { rr cc cc cc cc c l }
&& 프 & & 랑  && 스  && 의  & ~ &  \\
&& \textit{peu}& &  \textit{rang} &&  \textit{seu} && \textit{ui} & ~ & \textit{peurangseu-ui} (`{France}.\textsc{gen}') \\
{Eojeol}: && \multicolumn{8}{c}{\texttt{B-LOC}} & & \\ \cline{3-9}
{Morph}{eme}: && \multicolumn{6}{c}{\texttt{B-LOC}} & \texttt{O} & & \\  \cline{3-7}
{Syl}{la}{ble}: && \texttt{B-LOC} && \texttt{I-LOC} && \texttt{I-LOC} && \texttt{O} & & \\ \cline{3-3} \cline{5-5} \cline{7-7}
\end{tabular}
}
\caption{{Various approaches of annotation for named entities (NEs): the \textit{eojeol}-based approach annotates the entire word, the morpheme-based annotates only the morpheme and excludes the functional morphemes, and the syllable-based annotates syllable by syllable to exclude the functional morphemes.}}
\label{korean-ner-data-examples}
\end{figure}

{The morpheme-based approach has been adopted in the publicly available Korean Information Processing System (KIPS) corpora in 2016 and 2017. 
The two corpora} consist of 3660 and 3555 sentences respectively for training.\footnote{\url{https://github.com/KimByoungjae/klpNER2017}} 
Since the extraction targets of the morpheme-based NER are named entities on the morpheme level, {it requires a morphological analyzer to be used in the pre-processing step based on the input sentence.}
In 2016, \citet*{yu-ko:2016} proposed a morpheme-based NER system using a bidirectional LSTM-CRFs, in which they applied the morpheme-level Korean word embeddings to capture features for morpheme-based NEs on news corpora.
{Previous works such as \citet{lee-park-kim:2018} and \citet{kim-kim:2020} also used the same morpheme-based approach. In {these} studies, the morphological analysis model was integrated {into} NER.}
However, for the morpheme-based NER, considerable performance degradation occurs when converting {\textit{eojeols}} into morphemes. 
To mitigate this performance gap between morpheme- and {\textit{eojeol}}-based NER, \citet*{park-park-jang-EtAl:2017} proposed a hybrid NER model that combined {\textit{eojeol}-} and morpheme-based NER systems.
{The proposed \textit{eojeol-} and morpheme-based NER systems yield the best possible NER candidates based on an ensemble of both predicted results using bidirectional gated recurrent units (GRU) combined with conditional random fields (CRF)}.

In the last few years, {studies} on Korean NER have been focusing on \textit{eojeol}-based NER because such datasets {have become more widely available, including NAVER's} NER corpus which was originally prepared for a Korean NER competition in 2018.\footnote{\url{https://github.com/naver/nlp-challenge/tree/master/missions/ner}}
In \textit{eojeol}-based Korean NER tasks, the transformer-based model has been dominant because of its {outstanding} performance \citep*{min-na-shin-kim:2019,kim-oh-kim:2020,you-song-kim-yun-cheong:2021}.
Various transformer-based approaches, such as adding new features including the NE dictionary in a pre-processing procedure \citep*{kim-kim:2019}, {and} adapting byte-pair encoding (BPE) and morpheme-based features simultaneously \citep*{min-na-shin-kim:2019}, have been proposed.\label{bpe-abbr}
\citet*{min-na-shin-kim-kim:2021} proposed a novel {structure} for the Korean NER system that adapts the {architecture} of Machine Reading Comprehension (MRC) by extracting the start and end positions of the NEs from an input sentence.
{Although the performance of the MRC-based NER is inferior to those using token classification methods, it shows a possibility that the NER task can be solved based on the span detection problem.} 
Another notable approach for Korean NER is data argumentation.
\citet*{cho-kim:2021} achieved {improvement in the performance of the Korean NER model} using augmented data that replace tokens with the same NE category in the training corpus. 

To avoid potential issues of the \textit{eojeol}-based approach {where additional morphemes that should be excluded from NEs may be included}, syllable-based Korean NER corpus was proposed, {such as} in Korean language understanding evaluation (KLUE) in 2021 \citep{park-EtAl:2021:KLUE}. 
The {NE tags} in the corpus are annotated on the syllable level. {For instance, the NE tags of the word `경찰' (\textit{gyeongchal}, `police') are marked on the two syllables, namely 경 \textit{gyeong} bearing \texttt{B-ORG}, and 찰 \textit{chal} bearing \texttt{I-ORG}.}}
{Table~\ref{korean-ner-data} summarizes the currently available Korean NER datasets.}

\begin{table} [htbp]
\caption{{Summary of publicly available Korean NER datasets: KIPS = Korean Information Processing System (\texttt{DATE}, \texttt{LOC}, \texttt{ORG}, \texttt{PER}, \texttt{TIME});
NAVER (14 entity labels, which we will describe in the next section);
MODU (150 entity labels); 
KLUE = Korean Language Understanding Evaluation (KIPS's 5 labels, and \texttt{QUANTITY}); *KLUE's test dataset is only available through \url{https://klue-benchmark.com/}.} \label{korean-ner-data}}
{\begin{minipage}{0.85\textwidth}
\begin{tabular} {ll  ccc }\hline
Name & Type & Train & Dev & Test \\ \hline
2016 KIPS  & morpheme & 3,660 & - & - \\\hdashline
2017 KIPS  & morpheme & 3,555 & {501} & 2,569 \\\hdashline
NAVER NER   & eojoel & 90,000 & - & - \\\hdashline
MODU 2019 & syllable & {150,082} & - & - \\\hdashline
{MODU 2021} & syllable & {68,400} & {1,085} & {8,685} \\\hdashline
KLUE  & syllable & 21,008 & 5,000 & 5,000*\\
\hline
\end{tabular}
\end{minipage}}

\end{table}

{The morpheme-based approach in previous studies differs from the proposed method. In {traditional morpheme-based} approaches, only the sequences of morphemes are used, and there are no word boundaries displayed. {This is because of the nature of the annotated KIPS corpus in which the tokens included are merely morphemes.} 
Our proposed morpheme-based approach conforms to the current language annotation standard by using the CoNLL-U format and multiword annotation for the sequence of morphemes {in} a single eojeol, where morphemes and word boundaries are explicitly annotated. 
Limited studies have been conducted {on} the mechanism of efficiently annotating and recognizing NEs in Korean. 
{To tackle this problem, an annotation scheme for the Korean corpus by adopting the CoNLL-U format, which decomposes Korean words into morphemes and reduces the ambiguity of named entities in the original segmentation, is proposed in this study}. 
Morpheme-based segmentation for Korean has been proved beneficial. Many downstream applications for Korean language processing, such as POS tagging \citep*{jung-lee-hwang:2018:TALLIP,park-tyers:2019:LAW}, phrase-structure parsing \citep*{choi-park-choi:2012:SP-SEM-MRL,park-hong-cha:2016:PACLIC,kim-park:2022} and machine translation \citep*{park-hong-cha:2016:PACLIC,park-EtAl:2017:Cupral}, are based on the morpheme-based segmentation, in which all morphemes are separated {from each other}. 
{In these studies}, the morpheme-based segmentation is implemented to avoid data sparsity because the number of possible words in longer segmentation granularity (such as \textit{eojeols}) can be exponential {given} the characteristics {of Korean, an agglutinative language}. 
{Since a morpheme-based dependency parsing system for Universal Dependencies with better dependency parsing results has also been developed \citep{chen-EtAl:2022:COLING}, we are trying to create a consortium to develop the morphologically enhanced Universal Dependencies for other morphologically rich languages such as Basque, Finnish, French, German, Hungarian, Polish, and Swedish.} \label{previous-work-end}

\section{Representation of NEs for the Korean language} \label{conllu}

The current Universal Dependencies \citep{nivre-EtAl:2016:LREC,nivre-EtAl:2020:LREC} for Korean {uses} the tokenized word-based CoNLL-U format.
Addressing the NER problem using the annotation scheme of the Sejong corpus or other Korean language corpora is difficult because of the agglutinative characteristics of words in Korean.\footnote{Although the tokenized words in Korean as in the Penn Korean treebank \citep{han-EtAl:2002} and the \textit{eojeols} as {the word units} in the Sejong corpus are different, they basically use the punctuation mark tokenized \textit{eojeol} and the surface form \textit{eojeol}, respectively. Therefore, we distinguish them as word (or \textit{eojeol})-based {corpora} from the proposed morpheme-based annotation.}
They {adopt the \textit{eojeol}-based annotation} scheme which cannot handle {sequence-level morpheme boundaries of NEs because of the characteristics of agglutinative languages}. For example, an NE \textit{emmanuel unggaro} (\textsc{person}) without a nominative case marker instead of \textit{emmanuel unggaro-ga} (`Emanuel Ungaro-\textsc{nom}') should be extracted {for the purpose of NER}. 
However, this is not the case in previous work on NER for Korean.

We propose a novel approach for NEs in Korean by using morphologically separated words based on the morpheme-based CoNLL-U format of the Sejong POS tagged corpus proposed in \citet{park-tyers:2019:LAW}, which has successfully obtained better results in POS tagging {compared to using} the word-based Sejong corpus. While \citet{park-tyers:2019:LAW} have proposed the morpheme-based annotation scheme for POS tags and conceived the idea of using their scheme on NER, they have not proposed {any practical} method to adopt the annotation scheme to NER tasks. As a result, existing works have not explored {these} aspects such as how to fit the NE tags to the morpheme-based CoNLL-U scheme and how the NEs are represented in this format. Our proposed format for NER corpora considers the linguistic characteristics of Korean NE, and it also allows the automatic conversion between the word-based format and the morpheme-based format.\footnote{Details of the automatic conversion algorithm between the word-based format and the morpheme-based format {are addressed} in Section \ref{experiments}.}
Using the proposed annotation scheme in our work as demonstrated in Figure~\ref{sjtree-ner}, we can directly handle the word boundary problem between content and functional morphemes {while} using any sequence labeling algorithms.
{For example, only \textit{peurangseu} (`France') is annotated as a named entity (\texttt{B-LOC}) instead of \textit{peurangseu-ui} (`France-\textsc{gen}'), and 
\textit{emmanuel unggaro} (`Emanuel Ungaro-\') instead of \textit{emmanuel unggaro-ga} (`Emanuel Ungaro-\textsc{nom}') as \texttt{B-PER} and \texttt{I-PER} in Figure~\ref{sjtree-ner}. 
}

In the proposed annotation scheme, NEs are, therefore, no longer marked on the \textit{eojeol}-based
natural segmentation. Instead, each NE annotation corresponds to a specific set of morphemes that belong to the NE. Morphemes that do not belong to the NE are excluded from the set of the NE annotations and thus are not marked as part of the NE. As mentioned, this is achieved by adapting the CoNLL-U format as it provides morpheme-level segmentation of the Korean language. 
While those morphemes that do not belong to the NE are usually postpositions, determiners, and particles, this does not mean that all the postpositions, determiners, and particles are not able to be parts of the NE. An organization's name or the name of an artifact may include postpositions or the name of an artifact may include postpositions or particles in the middle, in which case they will not be excluded. The following NE illustrates the aforementioned case:\label{new-ne-example}

\begin{center}
{\small
\begin{tabular}{ l lll l  }
1& . . . & \multicolumn{3}{c}{} \\
2&`&\_& SpaceAfter=No&\\
3-4&프로페셔널의&\_&\_& \textit{peuropesyeoneol-ui}\\
3&프로페셔널&B-AFW & \_&\textit{peuropesyeoneol} (`professional')\\
4&의&I-AFW&\_&\textit{-ui} (`-\textsc{gen}')\\
5&원칙&I-AFW&SpaceAfter=No&\textit{wonchik} (`principle')\\
6&'&\_ & SpaceAfter=No&\\
7&은&\_&\_&\textit{-eun} (`-\textsc{top}')\\
8& . . . & \multicolumn{3}{c}{ } \\
\end{tabular}
}
\end{center} \label{professionla}

The NE 프로페셔널의 원칙 \textit{peuropesyeoneol-ui wonchik} (`the principle of professional') is the {title} of a book belonging to \texttt{AFW} (\textsc{artifacts/works}). Inside this NE, the genitive case marker \textit{-ui}, which is a particle, {remains a part of the NE.}
Because these exceptions can also be captured by sequence labeling, such an annotation scheme of Korean NER can provide a more detailed approach to NER tasks in which the NE annotation on \textit{eojeol}-based segments can now be decomposed to morpheme-based segments, and purposeless information can be excluded from NEs to improve the performance of the machine learning process.

\begin{figure}
\centering
\begin{tabular} {|p{0.9\textwidth}|}    \hline
{\small \# sent\_id = BTAA0001-00000012}\\
{\small \# text = 프랑스의 세계적인 의상 디자이너 엠마누엘 웅가로가 실내 장식용 직물 디자이너로 나섰다.} \\
\resizebox{.9\textwidth}{!}
{\small
\begin{tabular} {lll lll ll}
1-2&프랑스의&\_&\_&\_&\_  &\_  & \\
1&프랑스&프랑스&PROPN&NNP& B-LOC &\_  &\textit{peurangseu} (`France')\\
2&의&의&ADP&JKG&\_ &\_  & \textit{-ui} (`-\textsc{gen}')\\
3-6&세계적인&\_&\_&\_&\_ &\_  & \\
3&세계&세계&NOUN&NNG&\_&\_  & \textit{segye} (`world')\\
4&적&적&PART&XSN&\_&\_  &\textit{-jeok} (`-\textsc{suf}')\\
5&이&이&VERB&VCP&\_&\_  &\textit{-i} (`-\textsc{cop}')\\
6&ㄴ&은&PART&ETM&\_&\_  &\textit{-n} (`-\textsc{rel}')\\
7&의상&의상&NOUN&NNG&\_ &\_  & \textit{uisang} (`fashion')\\
8&디자이너&디자이너&NOUN&NNG&\_ &\_  & \textit{dijaineo} (`designer')\\
9&엠마누엘&엠마누엘&PROPN&NNP& B-PER &\_  & \textit{emmanuel} (`Emanuel')\\
10-11&웅가로가&\_&\_&\_&\_ &\_  &\\
10&웅가로&웅가로&PROPN&NNP& I-PER &\_  & \textit{unggaro} (`Ungaro')\\
11&가&가&ADP&JKS&\_ &\_  & \textit{-ga} (`-\textsc{nom}')\\
12&실내&실내&NOUN&NNG&\_ &\_  & \textit{silnae} (`interior')\\
13-14&장식용&\_&\_&\_&\_ &\_  &\\
13&장식&장식&NOUN&NNG&\_&\_  & \textit{jangsik} (`decoration')\\
14&용&용&PART&XSN&\_&\_  & \textit{-yong} (`usage')\\
15&직물&직물&NOUN&NNG&\_ &\_  & \textit{jikmul} (`textile')\\
16-17&디자이너로&\_&\_&\_&\_ &\_  &\\
16&디자이너&디자이너&NOUN&NNG&\_ &\_  &\textit{dijaineo} (`designer')\\
17&로&로&ADP&JKB&\_ &\_  & \textit{-ro} (`-\textsc{ajt}')\\
18-20&나섰다&\_&\_&\_&\_  &SpaceAfter=No &\\
18&나서&나서&VERB&VV&\_ &\_  & \textit{naseo} (`become')\\
19&었&었&PART&EP&\_ & \_  &\textit{-eoss}  (`\textsc{past}')\\
20&다&다&PART&EF&\_ &\_  & \textit{-da}  (`-\textsc{decl}')\\
21&.&.&PUNCT&SF&\_& \_  &\\
\end{tabular}} \\ \hline
\end{tabular}
\caption{CoNLL-U style annotation with multiword tokens for morphological analysis and POS tagging. It can include BIO-based NER annotation where B-LOC is for a beginning word {of} \textsc{location} and I-PER for an inside word of \textsc{person}.}
\label{sjtree-ner}
\end{figure}

\section{NER learning models} \label{learning}
In this section, we propose our baseline {and neural} models that consider the state-of-the-art feature representations. NER is considered to be a classification task that classifies possible NEs from a sentence, and the previously proposed NER models are mostly trained in a supervised manner. In conventional NER systems, the maximum entropy model \citep{chieu-ng:2003:CoNLL} or conditional random fields (CRFs) \citep{mccallum-li:2003:CoNLL} are used as classifiers. Recently, {neural network-based} models with continuous word representations outperformed the conventional models in lots of studies \citep{lample-EtAl:2016:NAACL,dernoncourt-lee-szolovits:2017:EMNLP,strubell-EtAl:2017:EMNLP,ghaddar-langlais:2018:COLING,jie-lu:2019:EMNLP-IJCNLP}. 
In this study, we mainly present {the neural network-based} systems.

\subsection{Baseline CRF{-based} model}
The objective  of the NER system is to predict the label sequence $Y=$ \{$y_1, y_2,.. , y_n$\}, given the sequence of words $X=$ \{$x_1, x_2,.. , x_n$\}, where $n$ denotes the number of words. As we presented in Section \ref{linguistics}, $y$ {is} the gold label of the standard NEs (PER, LOC, and ORG). An advantage of CRFs compared to previous sequence {labeling} algorithms, such as HMMs, is that their features can be {defined} as per our definition. We used binary tests for feature functions by distinguishing between unigram ($f_{y,x}$) and bigram ($f_{y',y,x}$) features as follows:
\begin{align*}
f_{y,x} (y_i, x_i) & = \mathbf{1}(y_i = y, x_i = x) \\
f_{y',y,x} (y_{i-1}, y_i, x_i) & = \mathbf{1}(y_{i-1} = y', y_i = y, x_i = x) 
\end{align*}
where $\mathbf{1}({condition})$ = 1 if the condition is satisfied and 0 otherwise. 
Here, $({condition})$ represents the input sequence $x$ at the current position $i$ with CRF label $y$. We used word and other linguistic information {such as POS labels} up to the $\pm$2-word context for each surrounding input sequence $x$ and their unigram and bigram features.\label{R3Q28}

\subsection{RNN-based model} \label{rnn}
There have been several proposed {neural} systems that adapt the long short-term memory (LSTM) neural networks \citep{lample-EtAl:2016:NAACL,strubell-EtAl:2017:EMNLP,jie-lu:2019:EMNLP-IJCNLP}. These systems consist of three different parts. First, the word representation layer that transforms the input sequence $X$ as a sequence of word representations.
Second, the encoder layer {encodes} the word representation into a contextualized representation using RNN-based {architectures}. 
Third, the classification layer that classifies NEs from the contextualized representation of each word.

We construct our baseline neural model {following} \citet{lample-EtAl:2016:NAACL} by using pre-trained {fastText} word embedding \citep{bojanowski-EtAl:2017:TACL,grave-EtAl:2018:LREC} as the word representation layer and LSTM as the encoder layer. The overall structure of our baseline neural model is displayed in Figure~\ref{fig:rnn-model}.

\begin{figure}
\centering
\includegraphics[width=0.9\linewidth]{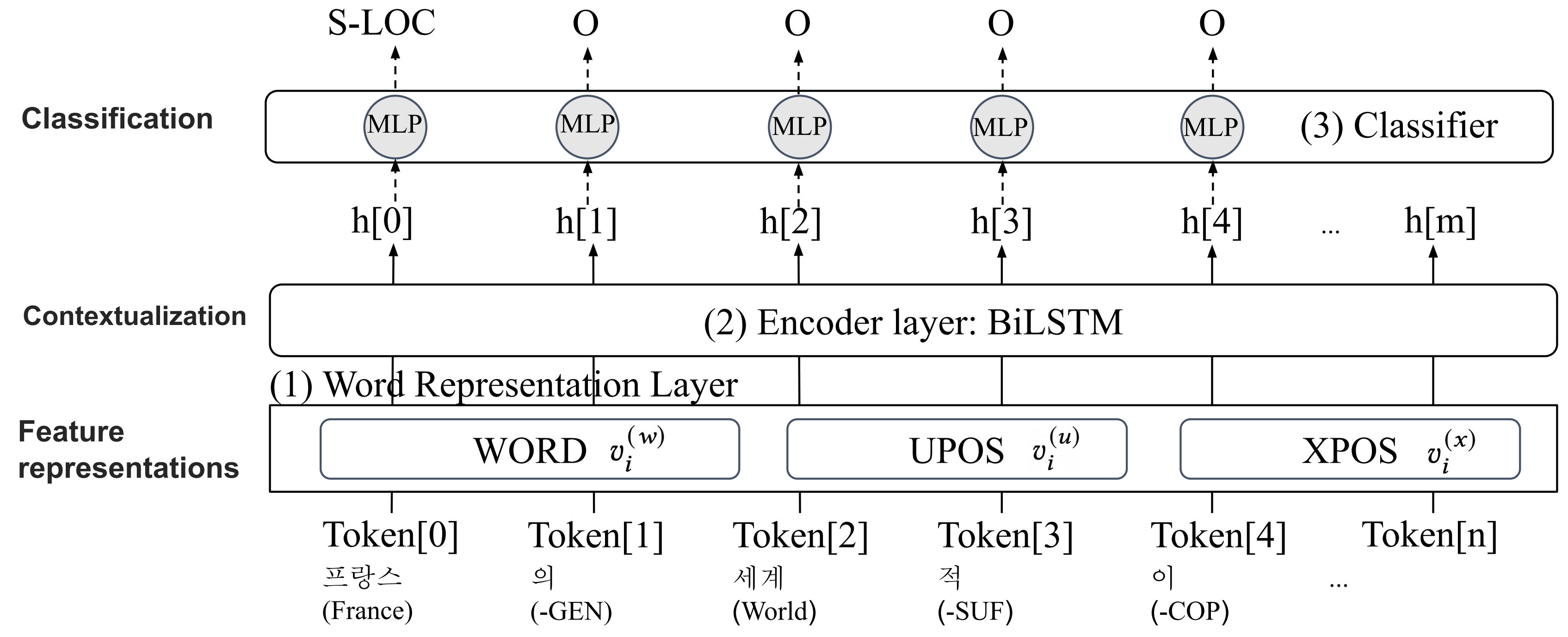}
\caption{Overall structure of {our} RNN-based model.}
\label{fig:rnn-model}
\end{figure}

\paragraph{Word representation layer} To train an {NER} system, the first step is to transform the words of each input sequence to the word vector {for the system to} learn the meaning of the words. 
{Word2vec \citep{mikolov-EtAl:2013:NIPS} and GloVe \citep{pennington-socher-manning:2014:EMNLP} are examples of word representations. Word2Vec is based on the distributional hypothesis and generates word vectors by utilizing co-occurrence frequencies within a corpus. FastText, on the other hand, is an extension of the Word2Vec that incorporates subword information. Rather than treating each word as a discrete entity, FastText breaks words down into smaller subword units, such as character n-grams, and learns embeddings. This approach allows FastText to capture information about word morphology and handle out-of-vocabulary (OOV) words more effectively than other methods.}

Given an input word $x_i$, we initialize word embedding $v^{(w)}_i$ using pre-trained fastText embeddings \citep{grave-EtAl:2018:LREC} provided by the 2018 CoNLL shared task organizer.
Randomly initialized POS features, UPOS $v^{(u)}_i$ and XPOS $v^{(x)}_i$ are concatenated to $v^{(w)}_i$ if available.

\paragraph{Encoder layer}
{The word embedding proposed in this study utilizes context-aware word vectors in a static environment. Therefore, in the NER domain we are using, it can be considered context-independent. For example,} each word representation $v^{(w)}_i$ {does not} consider contextual words such as $v^{(w)}_{i+1}$ and $v^{(w)}_{i-1}$. {Furthermore, since the POS features are also randomly initialized, there is an issue of not being able to determine the context.} LSTMs are typically used to address the context-independent problem. The sequence of word embeddings is fed to a two-layered LSTM to transform the representation to be contextualized as follows:

\begin{eqnarray}
h_i^{(ner)} = BiLSTM(r^{(ner)}_{0}, (v^{(w)}_{1}, .., v^{(w)}_{n}))_{i} 
\label{eq:bilstm}
\end{eqnarray}
\noindent where $r^{(ner)}_0$ represents the randomly initialized context vector, and $n$ denotes the number of word representations.

\paragraph{Classification layer}
The system consumes the contextualized vector $h^{(ner)}_i$ for predicting NEs using a multi-layer perceptron (MLP) classifier as follows:
\begin{eqnarray}
z_{i} = Q^{(ner)}MLP(h^{(ner)}_{i}) + b^{(ner)}
\label{eq:mlp}
\end{eqnarray}
\noindent where $MLP$ consists of a linear transformation with {the} ReLU function. 
The output $z_i$ is of size $1 \times k$, where $k$ denotes the number of distinct tags. 
We {apply} Softmax to convert the output, $z_i$, to be a distribution of probability for each NE.

During training, the system learns parameters $\theta$ that maximize the probability $P(y_{i}|x_{i},\theta)$ from the training set $T$ based on the conditional negative log-likelihood loss.
\begin{eqnarray}
 Loss(\theta) = & \sum_{(x_{i},y_{i})\in T}-\log P(y_{i}|x_{i},\theta) \\
 \hat{y}= & \arg\max_{y}P(y|x_{i},\theta)
\end{eqnarray}
where $\theta$ represents all the aforementioned parameters, including $Q^{(ner)}$, $MLP^{(ner)}$, $b^{(ner)}$, and $BiLSTM$. $(x_{i},y_{i})\in T$ denotes input data from training set $T$, and $y_{i}$ represents the gold label for input $x_{i}$. $\hat{y}=\arg\max_{y}P(y|x_{i},\theta)$ is a set of predicted labels. 

{It is common to replace the MLP-based classifiers with the CRF-based classifiers because the possible combinations of NEs are} not considered in MLP.
Following the baseline CRF model, we also implemented CRF models on top of LSTM output $h$ as the input of CRF. The performance of the CRFs and MLP {classifiers} is investigated in Section \ref{results}.

\subsection{BERT-based model} \label{BERT-base}

As mentioned in the previous section, {the word vectors we used were trained in a static environment, which led to a lack of context information.} To address this issue, a system should model the entire sentence as a source of contextual information. Consequently, the word representation method has been rapidly {adjusted} to adapt new embeddings trained by a masked language model (MLM). {The MLM randomly masks} some of the input tokens and predicts the masked word by restoring them to the original token. {During the prediction stage, the MLM} captures contextual information to make better predictions. 

{Recent studies show that} BERT trained by bidirectional transformers with {the} masked language model strategy achieved outperforming results in {many NLP tasks proposed in the General Language Understanding Evaluation
(GLUE) benchmark \citep{wang-etal-2018-glue}}. 
Since then, several {transformer-based} models have been proposed for NLP, and they show outstanding performances over almost all NLP tasks. In this section, we propose a NER system {that is based on the new annotation scheme and apply} the multilingual BERT and XLM-RoBERTa models to {the} Korean NER task. \label{new-annotation}

\paragraph{BERT representation layer}
{BERT's tokenizer uses a WordPiece-based approach, which breaks words into subword units. For example, the word ``snowing" is divided into ``snow" and ``\#\#ing". In contrast, the XLM-RoBERTa tokenizer uses a BPE-based 
approach \citep{sennrich-haddow-birch:2016:ACL}, which can produce more subwords than the WordPiece approach.} XLM-RoBERTa uses bidirectional modeling to check context information and uses BPE when training MLM. The XLM-RoBERTa decomposes a token  $x_i$ into a set of subwords. For example, a token 실내 \textit{silnae} (`interior') is decomposed to three subwords, namely  {`시 \textit{si}', `ㄹ \textit{l}', and `내 \textit{nae}'} by the XLM-RoBERTa tokenizer. 
{Since Korean has only a finite set of such subwords}, the problems of unknown words that do not appear in the training data $T$ are addressed effectively. 
Once the BERT representation is trained by MLM using massive plain texts, the word representation and all parameters {from} training are stored as the pre-trained BERT. Recently, there have been several pre-trained {transformer-based} models in public use. Users are generally able to fine-tune the pre-trained BERT representation for their downstream tasks. We implemented our BERT-based NER systems using the pre-trained BERT models provided by Hugging Face.\footnote{\url{https://github.com/huggingface/transformers}} 
{In the BERT-based system, a word is tokenized and transformed into a sequence of BERT representations as $V^{(b)}_{i} =$ \{$v^{(b)}_{i,1}, v^{(b)}_{i,2}, .., v^{(b)}_{i,j}$\}, where $i$ and $j$ denote the $i$-th word from the number of $n$ words and the number of BPE subwords of the $i$-th word, respectively.}
{We only take the first BPE subword for each word as the result of the BERT representations proposed in \cite{devlin-EtAl:2019:NAACL}. As an illustration, our system obtains embeddings for the subword ``snow" in the word ``snowing," which is segmented into ``snow" and ``\#\#ing" by its tokenizer.} This is because BERT is a bidirectional model, and even the first BPE subword for each token contains contextual information over the sentence. 
Finally, the resulting word representation becomes $V^{(b)} =$ \{$v^{(b)}_{1,1}, v^{(b)}_{2,1},.. , v^{(b)}_{n,1}$\}. 
Following {what has been done to} the RNN-based model, we concatenate the POS embedding to $v^{(b)}_{i,1}$ if available.

\paragraph{Classification layer}
We {apply} the same classification layer of the RNN-based model with an identical negative log-likelihood loss for the BERT-based model.

\section{Experiments} \label{experiments}

\subsection{Data} \label{experiments-data}
The Korean NER data we introduce in this study is from NAVER, which was originally prepared for a Korean NER competition in 2018.
NAVER’s data includes 90,000 sentences, and each sentence consists of indices, words/phrases, and NER annotation in the BIO-like format from the first column to the {following columns}, respectively. However, sentences in NAVER’s data were segmented based on \textit{eojeol}, which, as mentioned, is not the best way to represent Korean NEs.

We resegment all the sentences into the morpheme-based representation as described in \citet{park-tyers:2019:LAW}, such that the newly segmented sentences follow the CoNLL-U format. An {\texttt{eoj2morph}} script is implemented to map the NER annotation from {NAVER’s \textit{eojeol}-based data to the morpheme-based} data in the CoNLL-U format by pairing each character in NAVER’s data with the corresponding character in the CoNLL-U data, and removing particles and postpositions from the NEs mapped to the CoNLL-U data when necessary. {The following presents an example of such conversions:}

\begin{center}
{\small
\begin{tabular}{lll c lll l}
\multicolumn{3}{c}{\textsc{naver}} & & \multicolumn{3}{c}{(proposed) \textsc{conllu}} & \\
 &&    & & 1-2& 프랑스의& \_ & \textit{peurangseu-ui}\\
1 &프랑스의     &B-LOC& $\Rightarrow$ & 1& 프랑스& B-LOC &\textit{peurangseu} (`France')\\
&&&& 2& 의 & \_ & \textit{-ui} (`-\textsc{gen}')\\
\end{tabular}
}
\end{center}

In particular, the script includes several heuristics that determine {whether} the morphemes in an \textit{eojeol} belong to the named entity {the \textit{eojeol} refers to}. When both NAVER’s data that have \textit{eojeol}-based segmentation and the corresponding NE annotation, and the data {in the CoNLL-U format} that only {contain} morphemes, UPOS (universal POS labels, \citet{petrov-das-mcdonald:2012:LREC}) and XPOS (Sejong POS labels as language-specific POS labels) features, are provided, the script first {aligns} each \textit{eojeol} from NAVER’s data to its morphemes in the CoNLL-U data, and then determines the morphemes {in} this \textit{eojeol} that {should} carry the NE {tag(s)} of this \textit{eojeol}, if any.
The {criteria} {for} deciding {whether} the morphemes {are supposed to} carry the NE {tags} is that these morphemes should not be adpositions (prepositions and postpositions), punctuation marks, particles, determiners, or verbs. 
However, cases exist in which some \textit{eojeols} that carry {NEs} only contain morphemes of the aforementioned types. In {these cases}, when the script does not find any morpheme that can carry the NE {tag}, the size of the excluding POS set above will be reduced {for} the given \textit{eojeol}. The script will first attempt to find a verb in the \textit{eojeol} to bear the NE annotation, and subsequently, will attempt to find a particle or a determiner. The excluding set keeps shrinking until a morpheme is found to carry the NE annotation of the \textit{eojeol}.
Finally, the script marks the morphemes that are in between two NE {tags} 
{representing the same named entity}
as part of that {named entity} (e.g., a morpheme that is between B-LOC and I-LOC is marked as I-LOC), and assigns all other morphemes an ``O'' notation, where B, I, and O denote beginning, inside and outside, respectively. 
Because the official evaluation set is not publicly available, the {converted data in the CoNLL-U format from NAVER's training set} are then divided into three {subsets}, {namely} the training, holdout, and evaluation sets, with portions of 80\%, 10\%, and 10\%, respectively.
Algorithm~\ref{pseudo} describes the pseudo-code for converting {data from NAVER's \textit{eojeol}-based} format into the {morpheme-based} CoNLL-U format. 
Here, $w_i:t_i$ in $S_{\text{NAVER}}$ represents a word $w_i$ and its {NE tag} $t_i$ (either an entity label or \texttt{O}). 
$w'_{i-k} m_i ... m_k$ in $S_{\text{CoNLL-U}}$ displays a word $w'_{i-k}$ with its separated morphemes $m_i ... m_k$.

\begin{algorithm}[t]
\SetAlgoLined
\KwResult{CoNLL-U format for Korean NER}
align words $w$s between $S_{\text{NAVER}}$ and $S_{\text{CoNLL-U}}$
where $S_{\text{NAVER}} = {w_1:t_1 ... w_x:t_x}$ and $S_{\text{CoNLL-U}} = {w'_1 ... w'_{l-n}, m_l ... m_n, w'_{n+1} ... w'_y}$\;
\While{$w_i$ $S_{\text{NAVER}}$ and $w'_j$ $S_{\text{CoNLL-U}}$ where $i$ and $j$ are aligned indexes}{
\eIf{$w_i:t_i$ and $t_i$ == \texttt{O}}{
    assign \texttt{O} {to} $w'_j$ or \texttt{O}s {to} $m_j, ..., m_k$ for $w'_{j-k}$\;}
{
    \eIf{$w_i:t_i$ and $t_i$ != \texttt{O} and $w'_j$}
    {
    assign $t_i$ {to} $w'_j$
    }
    {
        \# in case of $w'_{j-k}$\;
        assign $t_i$ {to} $m_j$ \;
        \While{$m_h$ where $m_{j+1} .. m_{k}$}{
        assign I-$t_i$ until $m_{h}$ is not a functional morpheme\;
        }
    }
}
}
\caption{Pseudo-code for converting {data from NAVER’s \textit{eojeol}-based format into the morpheme-based CoNLL-U format}} \label{pseudo}
\end{algorithm}

{We also implement a \texttt{syl2morph} script that maps the NER annotation from syllable-based data (e.g., KLUE and MODU) to the data in the morpheme-based CoNLL-U format to further test our proposed scheme. 
While the \texttt{eoj2morph} script described above utilizes UPOS and XPOS tags to decide which morphemes in the \textit{eojeol} should carry the NE tags, they are not used in \texttt{syl2morph} anymore, as the NE tags annotated on the syllable level already excludes the syllables that belong to the functional morphemes. 
Additionally, NEs are tokenized as separate \textit{eojeols} at the first stage before being fed to the POS tagging model proposed by \citet{park-tyers:2019:LAW}. This is because the canonical forms of Korean morphemes do not always have the same surface representations as the syllables do. 
Because the \texttt{syl2morph} script basically follows the similar principles of \texttt{eoj2morph}, except for the two properties (or the lack thereof) mentioned above, we simply present an example of the conversion described above:} 

\begin{center}
{\small
\begin{tabular}{lll c lll l}
\multicolumn{3}{c}{\textsc{klue}} & & \multicolumn{3}{c}{(proposed) \textsc{conllu}} & \\
1&프&B-LOC& & 1-2& 프랑스의& \_ & \textit{peurangseu-ui}\\
2&랑&I-LOC& $\Rightarrow$ & 1& 프랑스& B-LOC &\textit{peurangseu} (`France')\\
3&스&I-LOC& & 2& 의 & \_ & \textit{-ui} (`-\textsc{gen}')\\
4&의&\_     & & &  &  & \\
\end{tabular}
}
\end{center}


{We further provide \texttt{morph2eoj} and \texttt{morph2syl} methods which allow back-conversion from the proposed morpheme-based format to either NAVER's \textit{eojeol}-based format or the syllable-based format, respectively. The alignment algorithm for back-conversion is simpler and more straightforward given that our proposed format preserves the original \textit{eojeol} segment at the top of the morphemes for each decomposed \textit{eojeol}. As a result, it is not necessary to align morphemes with \textit{eojeols} or syllables. Instead, only \textit{eojeol}-to-\textit{eojeol} or \textit{eojeol}-to-syllable matching is required.
The \texttt{morph2eoj} method assigns NE tags to the whole \textit{eojeol} that contains the morphemes these NE tags belong to, given that in the original \textit{eojeol}-based format, \textit{eojeols} are the minimal units to bear NE tags.} 
{The \texttt{morph2syl} method first locates the \textit{eojeol} that carries the NE tags in the same way as described above for \texttt{morph2eoj}. Based on the fact that NEs are tokenized as separate \textit{eojeols} by \texttt{syl2morph}, the script assigns the NE tags to each of the syllables in the \textit{eojeol}. 

{Back-conversions from the converted datasets in the proposed format to their original formats are performed. Both \textit{eojeol}-based and syllable-based datasets turned out to be identical to the original ones after going through conversions and back-conversions using the aforementioned script, which shows the effectiveness of the proposed algorithms. 
Manual inspection is conducted on parts of the converted data, and no error is found. While the converted dataset may contain some errors that manual inspection fails to discover, back-conversion as a stable inspection method recovers the datasets with no discrepancy. Therefore, we consider our conversion and back-conversion algorithms to be reliable. On the other hand, no further evaluation of the conversion script is conducted, mainly because the algorithms are tailored in a way that unlike machine learning approaches, linguistic features and rules of the Korean language, which are regular and stable, are employed during the conversion process.}

\subsection{Experimental setup}
Our feature set for baseline CRFs is described in Figure~\ref{crf-features}. 
We use {\texttt{crf++}\footnote{\url{https://taku910.github.io/crfpp}} {as} the implementation of CRFs, where \texttt{crf++}} automatically generates a set of feature functions using the template.  

\begin{figure}
\centering
{\small 
\begin{tabular}{|l l|} \hline
    \texttt{\# Unigram}&  \\
$w_{-2}$ & \\
$w_{-1}$ & \\
$w_{0}$ & (current word)\\
$w_{1}$ & \\
$w_{2}$ & \\
~& \\
\texttt{\# Bigram}& \\
$w_{-2}/w_{-1}$ & \\
$w_{-1}/w_{0}$ & \\
$w_{0}/w_{1}$ & \\
$w_{1}/w_{2}$ & \\\hline
\end{tabular}
}

\caption{CRF feature template example for the \textit{word} and \textit{named entity} data set.}
\label{crf-features}
\end{figure}

The hyperparameter settings for the neural models are listed in Table~\ref{tab:hyperparameter}{, Appendix \ref{appendix-hyperparameter}}. 
We apply the same hyperparameter settings as in \citet{lim-EtAl:2018:K18-2} for BiLSTM dimensions, MLP, optimizer including $\beta$, and learning rate to compare our results with those in a similar environment. We set 300 dimensions for the parameters {including} $LSTM$ in \eqref{eq:bilstm}, $Q$, and $MLP$ in \eqref{eq:mlp}. 
In the training phase, we train the {models} over the entire training dataset as an epoch with a batch size of 16. {For} each epoch, we evaluate the performance on the development set and save the best-performing model within 100 epochs.

The standard $F_{1}$ metric ($= 2 \cdot \frac{P \cdot R}{P + R}$) is used to evaluate NER systems, where precision ($P$) and recall ($R$) are as follows: 
\begin{align*} \label{eq}
P = \frac{\textrm{retrieved named entities}~\cap~\textrm{relevant named entities}}{\textrm{retrieved named entities}} \\
R  = \frac{\textrm{retrieved named entities}~\cap~\textrm{relevant named entities}}{\textrm{relevant named entities}}
\end{align*}
We evaluate our NER outputs on the official evaluation metric script provided by the organizer.\footnote{\url{https://github.com/naver/nlp-challenge}}

\section{Results} \label{results}

We {focus on the} following aspects of our results: (1) whether {the conversion of Korean NE corpora into the proposed morpheme-based CoNLL-U format} is more beneficial {compared to} the previously proposed \textit{eojeol}-based and syllable-based styles,
(2) the effect of multilingual {transformer-based} models, and (3) the impact of the additional POS features on Korean NER. 
{The outputs of the models trained using the proposed morpheme-based data {are converted back to their original format, either \textit{eojeol}-based or syllable-based}, before evaluation.}
{Subsequently, {all reported results are calculated in their original format for fair comparisons, given the fact that the numbers of tokens in different formats vary for the same sentence.} Nevertheless, it is worth noting that all experiments using the morpheme-based CoNLL-U data are both trained and predicted in this proposed format before conducting back-conversion.} \label{fair-comparison}

\subsection{Intrinsic results on various types of models} \label{intrinsic}

We compare the intrinsic results generated by different types of models. By saying ``intrinsic'', it implies that the results in this subsection differ owing to the ways and approaches of learning from the training data. We compare the performances of the baseline CRFs and our proposed neural models, and we also investigate the variations when our models use additional features in the data.

Table~\ref{bio+word+ne} summarizes the evaluation results on the test data based on the proposed machine learning models in Section \ref{learning}. 
Comparing the {transformer-based} models with LSTM+\textsc{crf}, {we found that} both mutlingual BERT-based models (\textsc{bert-multi}, \textsc{xlm-roberta}) outperformed LSTM+\textsc{crf}. {The} comparison reveals a clear trend: the word representation method is the most {effective} for the NER {tasks adopting our proposed scheme}. 
For the LSTM+\textsc{crf} model, we initialized its word representation, $v_{i}^{w}$, with a {\texttt{fastText}} word embedding \citep{joulin-EtAl:2016}. 
{However, once we initialized word representation using BERT, we observed performance improvements up to 3.4 points with the identical {neural network} structure {as shown in Table~\ref{bio+word+ne}}. Meanwhile, there are two notable observations in our experiment. The first observation is that the CRF classifier exhibits slightly better performance than the MLP {classifier}, with {an improvement of} 0.23, for \textsc{xlm-roberta}.}\label{R3Q16}
However, the improvement through the use of the CRF classifier is relatively {marginal} compared with the reported results of English NER \citep{ghaddar-langlais:2018:COLING}. 
{Moreover, when comparing the multilingual models (\textsc{bert-multilingual} and \textsc{xlm-roberta}) to the monolingual model (\textsc{klue-roberta}), we found that \textsc{klue-roberta} outperforms both \textsc{bert-multilingual} and \textsc{xlm-roberta}. This is because \textsc{klue-roberta} is trained solely on Korean texts and utilizes better tokenizers for the Korean language \citep{park-EtAl:2021:KLUE}.
}

\begin{table}
\caption{CRF/Neural results using different models {using NAVER's data converted into the proposed format}. \textsc{fastText} \citep{joulin-EtAl:2016} for \textsc{lstm+crf} word embeddings.} \label{bio+word+ne}
{\begin{minipage}{0.9\textwidth} 

\begin{tabular}{ c | c | cc | cc | cc} \hline
baseline & LSTM & \multicolumn{2}{c|}{\textsc{bert-multi}} & \multicolumn{2}{c|}{\textsc{xlm-roberta}} & \multicolumn{2}{c}{\textsc{klue-roberta}} \\ 
 \textsc{crf} & \textsc{+crf}  & \textsc{+mlp} & \textsc{+crf} & \textsc{+mlp} & \textsc{+crf} & \textsc{+mlp} & \textsc{+crf}  \\ \hline
{$71.50$} & {$84.76_{\pm0.29}$} & {$87.04_{\pm0.41}$} & {$87.28_{\pm0.37}$} & {$87.93_{\pm0.30}$} & {$88.16_{\pm0.33}$} & {$88.77_{\pm0.39}$} & {$88.84_{\pm0.43}$} \\ \hline
\end{tabular}
\end{minipage}}

\end{table}

Table~\ref{bio+word+upos+xpos+ne} details results using various sets of features. We use incremental words, UPOS, and XPOS, for the input sequence $x$ and their unigram and bigram features for CRFs. 
Both LSTM and XLM-RoBERTa achieve their best F$_{1}$ score when only the +UPOS feature is attached. 

\begin{table}
\caption{CRF/Neural results using the various sets of features {using NAVER's data converted into the proposed format}.
} \label{bio+word+upos+xpos+ne}
{\begin{minipage}{0.9\textwidth} 

\begin{tabular}{ c | cc | ccc } \hline
\multicolumn{1}{c|}{baseline \textsc{crf}} & \multicolumn{2}{c|}{\textsc{lstm+crf}} & \multicolumn{3}{c}{\textsc{xlm-roberta+crf}} \\ 
\textsc{word}  & \textsc{word} & \textsc{+upos} & \textsc{word} & \textsc{+upos} & \textsc{+xpos} \\ \hline
{$71.50$} & {$84.76_{\pm0.29}$} & {$84.94_{\pm0.34}$} & {$88.16_{\pm0.33}$} & \textbf{{$88.41_{\pm0.27}$}} & {$88.37_{\pm0.22}$}\\ \hline
\end{tabular}
\end{minipage}}
\end{table}

\subsection{Extrinsic results on different types of data}\label{result-extrinsic}

This subsection examines the extrinsic results given different types of data, whereas the previous subsection focuses on the differences in various models given only the {morpheme-based} CoNLL-U data. In this subsection, the performances of our models {trained on the datasets either in the proposed format or in their original formats,} and in either the BIO tagging format or the BIOES tagging format, are investigated.

As described in Table \ref{conllu+naver}, both the baseline CRF-based model and the BERT-based model achieve higher F$_{1}$ scores when the proposed CoNLL-U data are used, in contrast with NAVER’s \textit{eojeol}-based data. The testing data are organized {in a way} that the total number of tokens for evaluations remains the same, implying that the  F$_{1}$ scores generated by \texttt{conlleval} are fair for both groups. 
{This is realized by converting the model output of the morpheme-based CoNLL-U data back into NAVER's original \textit{eojeol}-based format such that they have the same number of tokens.} The {morpheme-based} CoNLL-U format outperforms the \textit{eojeol}-based format under the CRFs, whereas the CoNLL-U format still outperformed the \textit{eojeol}-based format by over 1\% {using} the BERT-based model.

Table~\ref{bio+bioes} presents the comparative results on two types of tagging formats where the models do not {use additional POS features}. {Previous studies show that} the BIOES annotation scheme {yields} superior results for several datasets {if the size of datasets is enough to disambiguate more number of labels} \citep{ratinov-roth:2009:CoNLL}.
Both the baseline CRF and neural models achieve higher F$_{1}$ scores than that of the BIOES tagging format when the BIO tagging format is used, which has two more types of labels -- E as endings, and S as single entity elements. {Our result reveals that adding more target labels} to the training data degrades model prediction {(14 labels $\times$ 2 for E and S)}. 
{Accordingly, we observe} that {in} this specific case, {adding} the two additional labels mentioned {increases} the difficulty of predictions.\label{R3Q35}

{Table \ref{modu+klue+klps} compares the results of the proposed morpheme-based CoNLL-U data and the syllable-based data. We use the \textsc{xlm-roberta+crf} model and only use word features for the experiments. Similar to the previous experiments, we back-convert the model output of the morpheme-based data back to the syllable-based format for fair comparisons. 
The results are consistent that for all four datasets, we observe performance improvement ranging from 3.12 to 4.31 points when the CoNLL-U format is adopted.}

{The performance difference between syllable-based NER results and morpheme-based NER results is mainly due to the fact that the \textsc{xlm-roberta+crf} model we used employs a subword-based tokenizer with larger units, rather than a syllable-based tokenizer. Therefore, one needs to use a BERT model with a syllable-based tokenizer for fair comparisons, if the goal is to only compare the performance between the morpheme-based format and the syllable-based format. However, this makes it difficult to conduct a fair performance evaluation in our study, because the evaluation we intended is based on BERT models trained in different environments when morpheme-based, \textit{eojeol}-based, or syllable-based data is given. Since subword-based tokenizers are widely employed in a lot of pre-trained models, our proposed format would benefit the syllable-based Korean NER corpora in a way that not only language models using syllable-based tokenizers can take them as the input, but those using BPE tokenizers or other types of subword-based tokenizers can also be trained on these syllable-based corpora once converted.}

\begin{table}
\caption{CRF/Neural result comparison between {the proposed} CoNLL-U {format} versus NAVER's {\textit{eojeol}-based format using NAVER's data}.} \label{conllu+naver}
{\begin{minipage}{0.9\textwidth} 
\begin{tabular}{ cc | cc } \hline
\multicolumn{2}{c|}{baseline \textsc{crf}} &  \multicolumn{2}{c}{\textsc{xlm-roberta+crf}} \\ 
\textsc{conllu} & \textsc{naver} & \textsc{conllu} & \textsc{naver}  \\ \hline
{$71.50$} & {$49.15$} & {$88.16_{\pm0.33}$} & {$86.72_{\pm0.49}$}\\ \hline
\end{tabular}
\end{minipage}}
\end{table}

\begin{table}
\caption{CRF/Neural result comparison between BIO versus BIOES annotations {using NAVER's data converted into the proposed format}.}  \label{bio+bioes}
{\begin{minipage}{0.9\textwidth} 
\begin{tabular}{ cc | cc } \hline
\multicolumn{2}{c|}{baseline \textsc{crf}} &  \multicolumn{2}{c}{\textsc{xlm-roberta+crf}} \\ 
\textsc{bio} & \textsc{bioes} & \textsc{bio} & \textsc{bioes}  \\ \hline
{$71.50$} & {$70.67$} & {$88.16_{\pm0.33}$} & {$85.70_{\pm0.35}$}\\ \hline
\end{tabular}
\end{minipage}}
\end{table}

\begin{table}
\caption{Result comparison between {the proposed} CoNLL-U {format} and {the syllable-based format using MODU (19 \& 21), KLUE, and ETRI datasets}. Model: \textsc{xlm-roberta+crf}}  \label{modu+klue+klps}
{\begin{minipage}{0.9\textwidth} 
\begin{tabular}{ cc | cc | cc | cc} \hline
\multicolumn{2}{c|}{MODU 19} &  \multicolumn{2}{c|}{MODU 21} & \multicolumn{2}{c|}{KLUE} & \multicolumn{2}{c}{ETRI}\\ 
\textsc{conllu} & \textsc{syllable} & \textsc{conllu} & \textsc{syllable} & \textsc{conllu} & \textsc{syllable} & \textsc{conllu} & \textsc{syllable} \\ \hline
{$88.03_{\pm0.20}$} & {$84.91_{\pm0.35}$} & {$81.72_{\pm0.31}$} & {$78.10_{\pm0.45}$} & {$91.72_{\pm0.29}$} & {$88.15_{\pm0.42}$} & {$97.59_{\pm0.12}$} & {$93.28_{\pm0.37}$}\\ \hline
\end{tabular}
\end{minipage}}
\end{table}




\subsection{POS Features and NER results}\label{pos-result}
\paragraph{Additional POS features help detect better B-$*$ tag sets} Comparing the XLM-RoBERTa models {trained with and without the UPOS tags, we observe an average performance improvement of 0.25\% when the UPOS feature is applied to the XLM-RoBERTa model}. However, if we only focus on the entities named {B-$*$}, the performance improvement increases {by 0.38}. We hypothesize that POS features in the NER system {play a role of evidence} in tagging decision-making for the start of its entities across tokens. 
For example, {a sentence in  \eqref{kharkiv-sent} exhibits  unknown words such as the names of a location: 하르키우와 돈바스 \textit{haleukiu-wa donbaseu} (`Kharkiv and Donbas').}

{
\begin{exe}
\ex \label{kharkiv}
\begin{xlist}
\ex \label{kharkiv-sent} \glll
하르키우와 돈바스 지역 방면을 향한 공세는 {~~} {...}\\
\textit{haleukiu-wa} \textit{donbaseu} \textit{jiyeog} \textit{bangmyeon-eul} \textit{hyanghan} \textit{gongseneun} {~~} ...\\
{Kharkiv-\textsc{conj}} {Donbas} {regions} {around-\textsc{acc}} {toward} {offensive-\textsc{top}} {~~} {...}\\
\trans `(The intensity of Russian) fire toward the regions of Kharkiv and Donbas ...'
\ex \label{kharkiv-pos} \begin{tabular}{lllll l}
1-2	& 하르키우와	&\_	&\_	&\_	 & \textit{haleukiu-wa} \\
1	&	하르키우	&	하르키우	&	PROPN	&	NNP & \textit{haleukiu} (`Kharkiv')\\
2	&	와	&	과	&	ADP	&	JC & \textit{-wa} (`\textsc{conj}')\\
3	&	돈바스	&	돈바스	&	PROPN	&	NNP & \textit{donbaseu} (`Donbas')\\
\multicolumn{5}{l}{...}
\end{tabular}
\end{xlist}
\end{exe}
}

{The NER system suffers from inadequate information, but when the system has POS information (proper noun, \texttt{PROPN} in UPOS and \texttt{NNP} in XPOS) as in \eqref{kharkiv-pos}, it can probably make {better decisions}. \label{unk}}

\paragraph{The UPOS tag set {benefits} Korean NER} \label{R3Q38}
Because {syntactic-relevant tasks} such as dependency parsing for Korean have mainly applied XPOS tag sets, we assumed the XPOS tags are more important than UPOS tags. {However, results in Table \ref{bio+word+upos+xpos+ne} demonstrate that applying the UPOS feature gives marginally better results than using the XPOS feature. {We consider this to be due to the predicted POS tags. Since the UPOS set contains 17 labels whereas the XPOS set (Sejong POS tag set) has 46 labels for Korean, the current prediction} model is more accurate in tagging UPOS labels than tagging XPOS labels. As a result, a more reliable UPOS feature provided by the tagging model also turns out to be more helpful in NER, compared with the XPOS feature.}

\subsection{Comparison between XLM-RoBERTa and BERT} \label{comparison-between}

\begin{figure}
\centering
\subfloat[BERT confusion matrix  \label{xpos-bert}]{\resizebox{.5\textwidth}{!}{{\includegraphics{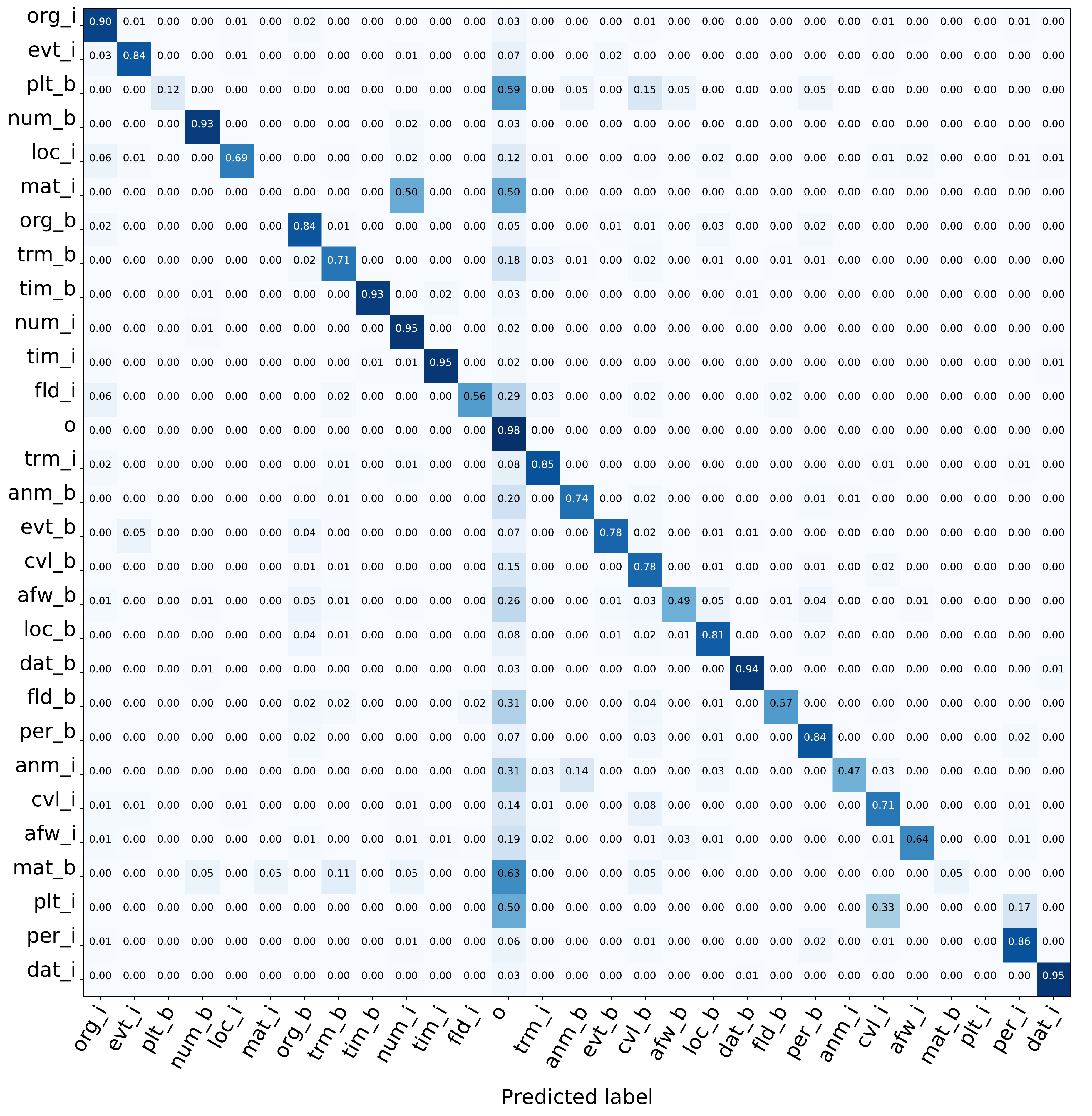}} }} ~~~~~~ \subfloat[XLM-RoBERTa confusion matrix \label{xpos-xlmroberta}]{\resizebox{.5\textwidth}{!}{{\includegraphics{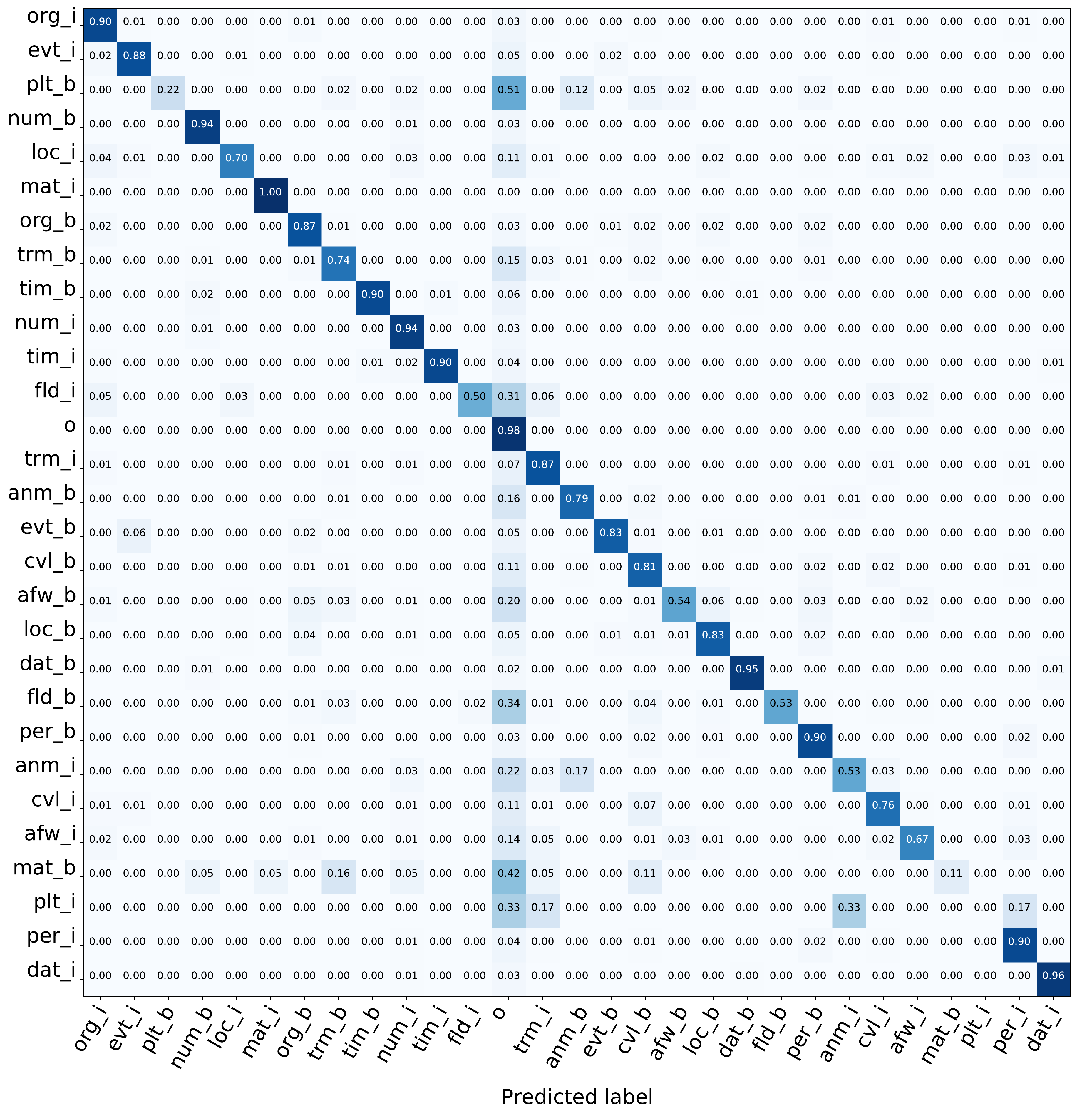}} }}

\caption{
Comparison of the confusion matrix between the BERT (\ref{xpos-bert}) and XLM-RoBERTa (\ref{xpos-xlmroberta}). Model on the test set. XLM-RoBERTa tends to show better results in finding the ``I-$*$'' typed entities, with +0.7\% of average score. 
}
\label{fig:chart}
\end{figure}

We reported that XLM-RoBERTa outperforms the multilingual BERT model in Table~\ref{bio+word+ne}. The differences between the two language models are discerned from Figure~\ref{fig:chart}, which displays the confusion matrix between the predicted and true labels for {both} BERT and XLM-RoBERTa. XLM-RoBERTa tends to exhibit superior results in finding the ``I-$*$'' typed entities, with +0.7\%  of {the} average score. Moreover, 
{it is observed that} the XLM-RoBERTa model exhibits fewer false predictions of \texttt{b-mat} and \texttt{i-plt} as \texttt{o} than the BERT model.

\section{Discussion}

{In this section, we discuss how the proposed scheme differs from the previously adopted schemes that represent NEs in Korean corpora.}
{As described {previously} in Figure~\ref{korean-ner-data-examples}, previous studies proposed other annotation formats for Korean NER,} {including \textit{eojeol}-based, morpheme-based, and syllable-based approaches.
We demonstrate how different annotation formats look like in Korean corpora in Figure~\ref{korean-ner-data-examples-files}, which also includes our proposed format.}

\begin{figure}
\centering
{\small
\begin{tabular}{| lll |lll |lll | lll |} \hline
\multicolumn{3}{|c|}{Eojeol} & \multicolumn{3}{c|}{Morpheme} & \multicolumn{3}{c|}{Syllable} & \multicolumn{3}{c|}{Proposed} \\\hline
1& 프랑스의 & \texttt{B-LOC} & 1& 프랑스 & \texttt{B-LOC} & 1& 프 & \texttt{B-LOC} & 1-2& 프랑스의 & \_\\
&&& 2 & 의 & \texttt{O} & 2& 랑 & \texttt{I-LOC} & 1 & 프랑스 & \texttt{B-LOC}\\
&&&&& & 3 & 스 & \texttt{I-LOC} & 2 & 의 & \texttt{O}\\
&&&&& & 4 & 의 & \texttt{O}     & & &\\ \hline
\end{tabular}}
\caption{{Various approaches of annotation for NEs in Korean}}
\label{korean-ner-data-examples-files}
\end{figure}

{
\subsection{Comparison with the eojeol-based format}
The \textit{eojeol}-based format, as adopted in NAVER's dataset, considers \textit{eojeols} as the tokens to which NE tags are assigned. The format preserves the natural segmentation of the Korean language, and is commonly used in other Korean processing tasks such as dependency parsing \citep{mcdonald-EtAl:2013:ACL,choi-EtAl:1994,chun-EtAl:2018:LREC}. On the other hand, taking \textit{eojeols} as the basic units for NER analyses has an inevitable defect. Korean is an agglutinative language in which functional morphemes, which are not parts of the NE, are attached to the \textit{eojeol} as well. As a result, it is impossible to exclude what does not belong to the NE from the token that bears the NE. 
Our proposed format has two advantages compared to the \textit{eojeol}-based format. While the functional morphemes cannot be excluded from the NE annotation, a morpheme-based format deals with this problem properly, as the tokens are no longer \textit{eojeols}, but morphemes instead. Consequently, the NE tags are assigned to the morphemes that are not functional. Moreover, since our proposed format keeps the \textit{eojeol} representation on the top of its morphemes, the information on the natural segmentation of the sentence is still preserved. The advantages of the proposed scheme are testified through experiments, and the results presented in Table \ref{conllu+naver}, Section \ref{result-extrinsic} show that the format in which functional morphemes are excluded from NE labels improves the performance of the NER models trained on the same set of sentences but in different formats. 
}

{
\subsection{Comparison with the previous morpheme-based format}
Previous studies on Korean NER attempted to employ morpheme-based annotation formats to tackle the problem of Korean's natural segmentation which is not able to exclude functional morphemes. As discussed in Section \ref{previous-work}, the KIPS corpora consider morphemes to be the basic units for NER analyses, and therefore the NE tags are assigned to only those morphemes that belong to the NE. While the morpheme-based format adopted in the KIPS corpora seems to be able to solve the problem the \textit{eojeol}-based format has, it also introduces new problems. The information on the natural segmentation (i.e., \textit{eojeols}) has been completely lost, and it is no longer possible to tell which morphemes belong to which \textit{eojeol}. 
This poses challenges both for human researchers to understand the corpora, and for the conversion from the morpheme-based format back to the \textit{eojeol}-based format. Not being able to observe the word boundaries and how the morphemes constitute \textit{eojeols} makes it hard for linguists and computer scientists to understand the language documented in the corpora. Moreover, since there is no information that suggests which morphemes should form an \textit{eojeol}, the script is not able to recover the corresponding \textit{eojeol}-based representation with NE tags. 
Our proposed format preserves all the \textit{eojeols} in the dataset, and clearly states which morphemes belong to the \textit{eojeols} through indices, as illustrated in Figure \ref{korean-ner-data-examples-files}. Conversions back to the \textit{eojeol}-based format has been made possible because of the \textit{eojeol}-level representations, which the \texttt{morph2eoj} script can refer to. Meanwhile, the proposed format is much more human-readable and presents linguistically richer information, which benefits the language documentation of Korean as well. 
}

{
\subsection{Comparison with the syllable-based format}
Another annotation format that deals with the defect of the \textit{eojeol}-based format, in which functional morphemes cannot be excluded, is the syllable-based format. Given that any syllable in Korean is never separated or shared by two morphemes or \textit{eojeols}, marking the NE tags on those syllables which do not belong to functional morphemes solves the problem as well. However, there are some potential issues when adopting this straightforward approach. The major problem here is that most, if not all, linguistic properties are lost when the sentences are decomposed into syllables, as a syllable is an even smaller unit than a morpheme which is the smallest constituent that can bear meaning. Moreover, it is impossible to assign POS tags to the syllable-based data, which have been proven helpful for Korean NER using morpheme-based data as presented in Table \ref{bio+word+upos+xpos+ne}. It is also not ideal to train a model which does not employ a syllable-level tokenizer using syllable-based corpora as discussed in Section \ref{result-extrinsic}, whereas many transformer-based models adopt BPE tokenization.  
Our proposed format addresses the above issues by preserving linguistic information and properties in the smallest units possible, namely morphemes. POS tags can therefore be assigned to each token as long as it carries meaning. 
We also provide a script that performs conversion from the syllable-based dataset into a morpheme-based dataset, such that a model using a subword-based tokenizer can fully utilize syllable-based corpora as additional data for further training. 
}

\section{Conclusion} \label{conclusion}
In this study, we {leverage the morpheme-based CoNLL-U format based on \citet{park-tyers:2019:LAW} and propose a new scheme that is compatible with Korean NER corpora}. Compared with the conventional \textit{eojeol}-based format in which the real pieces of NEs are not distinguished from the postpositions or particles inside the \textit{eojeol}, the proposed CoNLL-U format {represents} NEs on {the} morpheme-based level such that NE annotations are assigned to the exact segments representing the entities. The results of both the CRF models and the neural models reveal that our {morpheme-based} format outperforms {both }the \textit{eojeol}-based format {and the syllable-based format currently adopted by various Korean NER corpora}. We also examine the performance of {the above} formats {using} various additional features and tagging formats (BIO and BIOES). Meanwhile, the paper also investigates the linguistic features of the Korean language that are related to NER. This includes the distribution of postpositions and particles after NEs, the writing system of Korean, and the \textit{eojeol}-based segmentation of Korean. The linguistic features we stated further justify the annotation scheme that we propose. Finally, we {confirm} the functionality of our {NER} models, and the results obtained reveal the differences among all these models, where the XLM-RoBERTa model outperformed other types of models in our study. The results of the study validate the feasibility of the proposed approach to {the} recognition of {named entities in Korean}, which is justified to be appropriate linguistically. The proposed method of recognizing NEs can be used in several real-world applications such as text summarization, document search and retrieval, and machine translation. 


\appendix

\section{Hyperparameters of the Neural Models}\label{appendix-hyperparameter}

\begin{table}[ht]
{\begin{minipage}{0.9\textwidth} 
  \caption{Hyperparameters.}   \label{tab:hyperparameter}
    \begin{tabular}{l | c } \hline
{component} & {value} \\ \hline
 $Q$ (parameters) dim. & 300 \\ \hdashline
 $v_i^{(w)}$ (fastText) dim. & 300 \\ \hdashline
$v_i^{(u)}$ (UPOS) dim. & 50 \\ \hdashline
$v_i^{(x)}$ (XPOS) dim. & 50 \\ \hdashline
 $v_{i,j}^{(b)}$ (BERT) dim. & 768 \\ \hdashline
 BiLSTM hidden dim. & 512 \\ \hdashline
 No. BiLSTM layers & 2 \\ \hdashline
 MLP output dim. & 300 \\ \hdashline
 Dropout & 0.3 \\ \hdashline
 Learning rate of RNN & 0.002  \\ \hdashline
 Learning rate of BERT & 0.00002  \\ \hdashline
 $\beta_1$, $\beta_2$  & 0.9, 0.99  \\ \hdashline
 Epoch & 100  \\ \hdashline
 Batch size  & 16  \\ \hdashline
 {Random Seed} & 42  \\ \hdashline
 Gradient clipping & 5.0 \\ \hline
  \end{tabular}
\end{minipage}}
\end{table}

\end{document}